\documentclass[10pt,twocolumn,letterpaper]{article}

\usepackage[pagenumbers]{cvpr}

\usepackage{graphicx}
\usepackage{amsmath}
\usepackage{amssymb}
\usepackage{booktabs}

\usepackage[accsupp]{axessibility}

\usepackage{amsmath,amsfonts,bm}

\def\eqref#1{equation~\ref{#1}}

\def\1{\bm{1}}

\def\rc{{\textnormal{c}}}

\def\rw{{\textnormal{w}}}
\def\rx{{\textnormal{x}}}

\def\rz{{\textnormal{z}}}

\def\rvz{{\mathbf{z}}}

\DeclareMathAlphabet{\mathsfit}{\encodingdefault}{\sfdefault}{m}{sl}
\SetMathAlphabet{\mathsfit}{bold}{\encodingdefault}{\sfdefault}{bx}{n}

\newcommand{\E}{\mathbb{E}}

\newcommand{\KL}{D_{\mathrm{KL}}}

\usepackage{microtype}

\usepackage{subcaption}
\usepackage{textcomp}
\makeatletter
\@namedef{ver@everyshi.sty}{}
\makeatother
\usepackage{tikz}
\usetikzlibrary{fit,positioning}

\newcommand{\myparagraph}[1]{\vspace{2pt}\noindent{\bf #1}}

\usepackage[pagebackref,breaklinks,colorlinks]{hyperref}

\usepackage[capitalize]{cleveref}
\crefname{section}{Sec.}{Secs.}
\Crefname{section}{Section}{Sections}
\Crefname{table}{Table}{Tables}
\crefname{table}{Tab.}{Tabs.}

\def\confName{CVPR}
\def\confYear{2022}

\begin{document}

\title{Compositional Mixture Representations for Vision and Text}

\author{Stephan~Alaniz\textsuperscript{1,2} \hspace{2mm}
Marco~Federici\textsuperscript{3} \hspace{2mm}
Zeynep~Akata\textsuperscript{1,2}\\
\textsuperscript{1}University of Tübingen \hspace{2mm}
\textsuperscript{2}Max Planck Institute for Informatics \hspace{2mm}
\textsuperscript{3}University of Amsterdam\\
{\tt\small \{stephan.alaniz,zeynep.akata\}@uni-tuebingen.de, m.federici@uva.nl}}
\maketitle


\begin{abstract}
Learning a common representation space between vision and language allows deep networks to relate objects in the image to the corresponding semantic meaning. We present a model that learns a shared Gaussian mixture representation imposing the compositionality of the text onto the visual domain without having explicit location supervision. By combining the spatial transformer with a representation learning approach we learn to split images into separately encoded patches to associate visual and textual representations in an interpretable manner. On variations of MNIST and CIFAR10, our model is able to perform weakly supervised object detection and demonstrates its ability to extrapolate to unseen combination of objects.
\end{abstract}

\section{Introduction}
For an artificial intelligence agent to gain an understanding of the world comparable to the one of humans', it should to be able to connect the visual world with its semantic meaning. There has been a substantial effort in learning image representations \cite{kingma2014, Chen16} as well as text representation \cite{Mikolov13, Pennington14, DevlinCLT19}, capturing the semantic meaning and disentangling the variation factors in a way similar to how humans learn their surroundings. However, learning unsupervised visual representations from the image data, (e.g., through image reconstruction), can be challenging because there is no guidance towards an informative content (e.g., presence of objects) and uninformative content (e.g., the exact pixel location of the horizon), when there is no supervision involved or the downstream task is unknown \cite{Zhao17}. \cite{Bengio17} hypothesises that language can be a good prior towards forming useful representations, i.e., things that are commonly described or talked about by people in visual data is information we would like to preserve in our representations.

Our goal is to build upon the idea of preserving the semantic information in the learned representations and learn an unified representation between the text and images, where we use the semantic structure of the text to disentangle the visual data. For instance, when a caption mentions a car, the corresponding representation of the image should not only include the same information, but should be able to report as well, which part of the image caused the representation of "car". Naturally, such a representation is not limited to hold only a single concept, but can be composed of several individual components, both on the text as well as on the image side. As text is already highly structural, its representation can be viewed as the aggregation of its words' meanings. Images could be then decomposed into patches associated with the building blocks that encode the semantics within the text.

By directly modeling the compositionality of the representation, it becomes possible to obtain several desirable properties. The first property is generalizability. In other words, learning a representation of its constituents is usually simpler than inferring a meaningful joint representation of a complex example. Hence, in novel scenarios, being able to robustly embed partial entities, which are well known from the training set, allow representations to generalize more easily.  The second property is combinatorial extrapolation. Certain objects in the images might have high co-occurrence probability. Previously unknown object combinations during test time can cause a deep model to fail. Compositionality can overcome this bias as it allows to arbitrarily combine partial representations. The third and the final property is interpretability. Associating the visual components of the image to the semantics encoded by the textual components allows humans to develop a better understanding of how deep models form their complex representations and might also assist in identifying the causes behind possible failures.

\begin{figure*}
\centering
\begin{subfigure}{.8\textwidth}
  \centering
  \includegraphics[width=\linewidth]{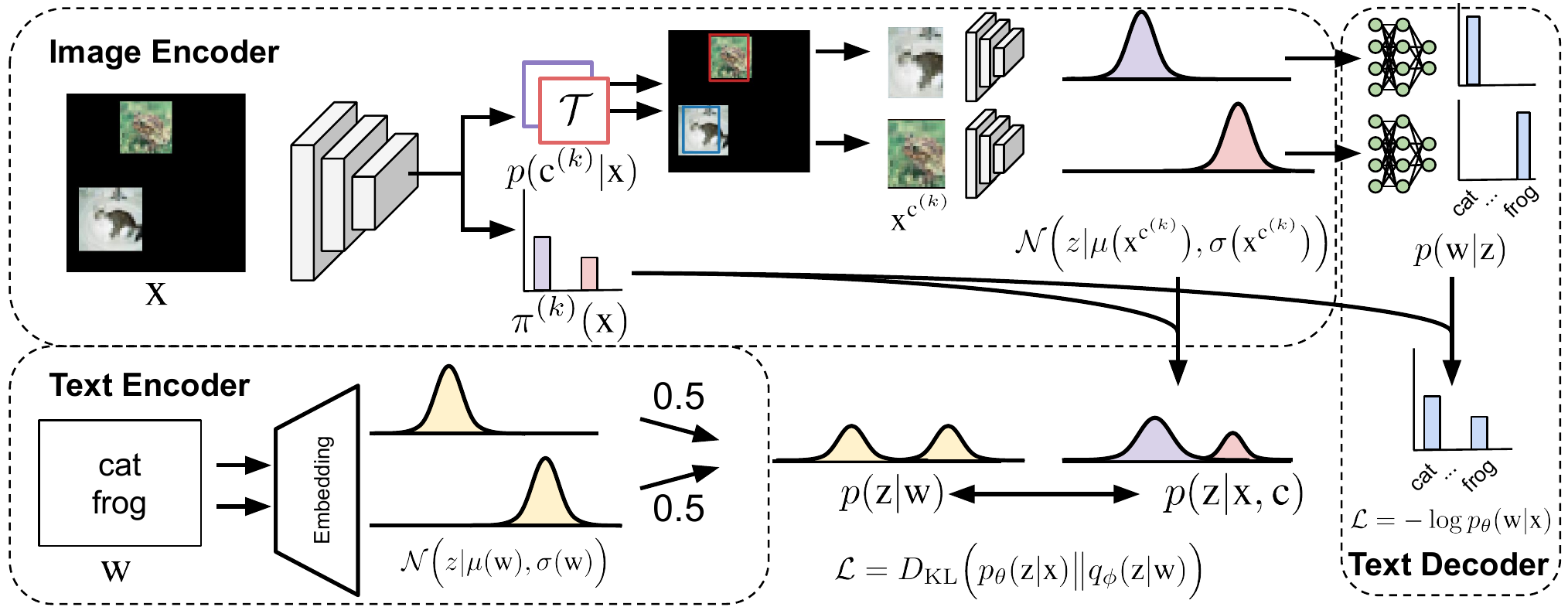}
  \caption{}
  \label{fig:arch}
\end{subfigure}%
\hfill
\begin{subfigure}{.19\textwidth}
  \centering
  \begin{tikzpicture}
    \tikzstyle{main}=[circle, minimum size = 8mm, thick, draw =black!80, node distance = 6mm]
    \tikzstyle{connect}=[-latex, thick]
    \tikzstyle{tie}=[dotted, thick]
    \tikzstyle{box}=[rectangle, draw=black!100]
      \node[main, fill = white!100] (x) {$\rx$};
      \node[main] (c) [below=of x] {$\rc^{(k)}$};
      \node[main] (z) [below=of c] {$\rz$};
      \node[main] (wt) [below=of z] {$\rw_i$};
      \path (x) edge [connect] (c)
            (x) edge [connect, bend right] (z)
            (c) edge [connect] (z)
            (z) edge [connect] (wt);
            
      \node[rectangle, inner sep=-2mm, fit= (c),label=below right:{K}, xshift=0mm] {};
      \node[rectangle, inner sep=3mm, draw=black!100, fit= (c)] {};
      \node[rectangle, inner sep=-2mm, fit= (wt),label=below right:{I}, xshift=1mm] {};
      \node[rectangle, inner sep=3mm,draw=black!100, fit= (wt)] {};
    
  \end{tikzpicture}
  \caption{}
  \label{fig:graph}
\end{subfigure}
\caption{Compositional Mixture (CoMix) Model. The graphical model is shown in the (b) while (a) shows the architecture of our proposed CoMix model that learns two Gaussian mixtures, one is learned from the image and the other one is from the text data. The image encoder first uses a CNN to predict the Gaussian mixture weights $\pi^{(k)}(\rx)$ as well as the transformation parameters $p(\rc^{(k)}|\rx)$ used by the spatial transformer module to extract image patches $\rx^{c^{(k)}}$. Each patch is individually encoded by a second CNN into Gaussian distributions $\mathcal{N}(z|\mu(\rx^{c^{(k)}}),\sigma(\rx^{c^{(k)}}))$. A text decoder $p(\rw|\rz)$ is learned with a negative log likelihood loss and ensures that textual information is contained in the representation.
A text encoder embeds individual words into Gaussian components and then mixes them into the textual Gaussian mixture representation $p(\rz|\rw)$. A KL-divergence loss allow to learn the correspondence between text tokens and image crops by matching the two representations. Without any bounding box supervision, our CoMix model learns to detect images.}
\end{figure*}
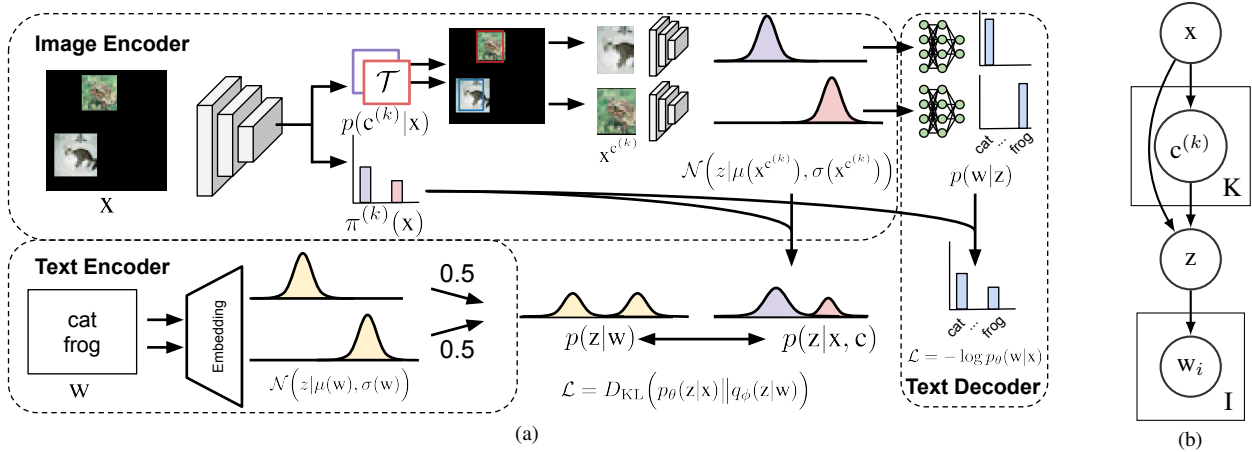

In this work, we implement the idea of using textual guidance to learn compositional image and text representations. One goal of the representation learning is to create representations of the raw input data, which are useful for an a priori unknown downstream tasks. We choose to evaluate our approach on image retrieval and weakly supervised object detection. The latter task is particularly fitting as an aim to decompose images into their respective objects as given by the textual labels without having any bounding box supervision. Therefore, it allows to give a good indication about the compositionality and disentanglement of the image representation that is achieved by our model.

Our contributions are as follows: we present a novel model that learns a compositional Gaussian mixture representation for both the image and the text and matches them with a KL divergence loss. The vision part of the model incorporates the spatial-transformer architecture to learn bounding boxes of the image parts corresponding to the different textual components. We evaluate our model on the altered MNIST and CIFAR10 datasets, where each image is composed of several images from the original dataset.

\section{Related work}

\myparagraph{Representation Learning.} We distinguish ourselves from most representation learning approaches on images in that we do not try to embed the complete image information in the latent code. Nonetheless, the goals of representation learning are shared across related work in this direction. MONet~\cite{Burgess19} also employs a compositional approach to scene understanding and reconstruction. It iteratively constructs a scene by its components such as objects and background elements and uses a VAE~\cite{kingma2014} architecture at each step. Similarly, IODINE~\cite{GreffKKWBZMBL19} also employs a VAE iteratively to infer and reconstructs scene an object at a time.
Related work has previously combined the spatial transformer architecture~\cite{Jaderberg15} with image representation learning both iteratively~\cite{Eslami16} and by processing the whole image at once~\cite{Crawford19}.

\myparagraph{Multi-modal Learning.} When considering models working with multi-modal data, MVAE~\cite{Wu18} applies the VAE setting to multiple modalities such as image and text and can be trained semi-supervised, but it does not employ a compositionality approach. Learning shared representations between images and text is often done targeting a specific application such as retrieval~\cite{Zheng17} or grounding sentences in images~\cite{RohrbachRHDS16}. Work on associating objects with parts of language through visual-semantic alignment~\cite{Karpathy14, Karpathy15} usually uses greater level of supervision at least on object detection. One such example is VisualBERT~\cite{Harold19} which combines the rich textual representations of BERT~\cite{DevlinCLT19} with supervised object detectors to solve a variety of visual-language tasks. Similarly, \cite{LiangZY20} uses natural language explanations to improve the performance of a visual classifier. LXMERT~\cite{TanB19} and UNITER~\cite{ChenLYK0G0020} demonstrate that rich representations learned from multi-modal vision and text data can benefit diverse tasks such as Visual Question Answering (VQA). The learned representations of CLIP~\cite{RadfordKHRGASAM21} achieve zero-shot object recognition capabilities by training on a large dataset of unstructured image-text pairs. Recently, several works~\cite{JiaYXCPPLSLD21,Huo2103,Shen2107,Li2110} build upon the contrastive learning objective similar to CLIP to learned multi-modal representations in an unsupervised manner.

\myparagraph{Weakly Supervised Object Detection.} Following the advances in supervised object detection in models such as YOLO~\cite{RedmonDGF16}, Faster-RCNN~\cite{RenHGS15} and Mask-RCNN~\cite{HeGDG17}, more work is dedicated in solving the object detection task without bounding box labels, i.e. weakly supervised with only labels about which object are present in the image. WSDDN~\cite{BilenV16} works by pooling spatial regions on the last convolutional feature layer of a CNN and has been challenged by similar approaches such as C-MIDN~\cite{Gao_2019_ICCV} and PredNet~\cite{ArunJK19}. These approaches, however, do not offer the same level of introspection and interpretability as our CoMix.

\section{Compositional Mixture Model}
We present our proposed compositional mixture model (CoMix) in the following section. In our setting, we consider a joint data distribution $p(\rx, \rw)$ of images $\rx$ and text $\rw$ with vocabulary $\mathcal{V}$. Our goal is to use a deep learning model to encode the visual signal coming from the images into a compositional representation, which entails all the semantic information that the components of the textual counterpart contains. At the same time, the representation should be interpretable in terms of which image region contributes to its components' representations.

\subsection{CoMix Overview}
In our CoMix model, we choose to model the latent representation as a Gaussian mixture. By being a multimodal distribution, it is able to model complex data while having the desired property of being compositional as it consists of several simple Gaussian distributions. Hence, each textual component as well as inferred image patches are embedded into individual Gaussians before being mixed to form the final representation.
Our CoMix model consists of three parts as depicted by Figure~\ref{fig:arch}. A text encoder $p(\rz|\rw)$ takes as input the text tokens $\rw$ and encodes each one of them into a latent mixture component.

Analogously, an image encoder maps the input image to another Gaussian mixture $p(\rz|\rx)$, where each Gaussian component corresponds to a different spatial image region. By aligning the two mixture distributions with a KL-divergence loss term, CoMix learns to associate components of the text to the concrete image regions. Attending to separate image patches is learned end-to-end by a spatial transformer module without requiring any extra supervision. Finally, a text decoder $p(\rw|\rz)$ ensures that the learned Gaussian mixture representation retains all the textual information, effectively preventing degenerate solutions. Thus, CoMix is trained by supplying image-text pairs $(\rx, \rw)$ without any additional supervision as to how these two modalities are related, e.g. there is no bounding box supervision on where the text is grounded in the image.

\subsection{Text Encoder}
A string of text $\rw$, such as a sentence, is composed of entities of interest, such as objects, that are also present in the image. Each word is embedded into a Gaussian latent component $\mathcal{N}(z|\mu(\rw_i), \sigma(\rw_i))$, where $\mu(\rw_i)$ and $\sigma(\rw_i)$ are determined by an embedding layer. The latent variables of all the words of the text are then combined into a Gaussian mixture model
\begin{equation}
    \frac{1}{|\rw|}\sum_i\mathcal{N}\Big(z|\mu(\rw_i), \sigma(\rw_i)\Big),
\end{equation} 
where each Gaussian of the mixture is weighted equally. This allows us to express a complex multi-modal representation as the sum of simple uni-modal Gaussian components representing its constituent words. Given a one-hot vector for each word, the text encoder is a single matrix that stores a trainable set of Gaussian parameters for each word.

\subsection{Image Encoder}
A spatial transformer module~\cite{Jaderberg15} determines a total of $k$ image regions, which are embedded into the parameters of a Gaussian distribution. 
Such an architecture allows us to crop, translate, and scale portions of the original input image by defining an affine transformation $A_{\theta}$ that maps image pixel locations of the source and the target image:

\begin{equation}
    \begin{pmatrix}
        x^s_i\\
        y^s_i
    \end{pmatrix} =
    A_{\theta}
    \begin{pmatrix}
        x^t_i\\
        y^t_i\\
        1
    \end{pmatrix} 
\end{equation}

where $(x^s, y^s)$ and $(x^t, y^t)$ are the source and target coordinates 
of pixel locations. 
The spatial transformer can be viewed as hard attention on the input image with differentiable transformation parameters $A_{\theta}$.

We employ a convolutional neural network (CNN) to learn $k$ different transformations $p(\rc^{(k)}|\rx)$ from the input image. By applying the spatial transformer on each of these, we get $k$ different image crops $\rx^{c^{(k)}}$, which are then individually encoded with another CNN that, in its turn, provides the parameters $\mu$ and $\sigma$ to a Gaussian distribution $\mathcal{N}(z|\mu(\rx^{c^{(k)}}), \sigma(\rx^{c^{(k)}}))$.
Hence, each Gaussian component of the image is also associated to an image region by a hard-attention mechanism. The Gaussian components are combined through the categorical mixture distribution $\pi^{(k)}(\rx)$, which is anticipated from the original input image with the same network that predicts the $k$ transformations $\rc^{(k)}$ to form the Gaussian mixture
$\sum_k\pi^{(k)}\mathcal{N}(z|\mu(\rx^{c^{(k)}}),\sigma(\rx^{c^{(k)}}))$.

\subsection{Representation Learning}
Our requirements for learning a meaningful representation are twofold: 1) The data likelihood under our graphical model needs to be maximized. Since the prediction of $\rw$ depends only on $\rz$, this forces $\rz$ to capture all the required textual information from the corresponding images $\rx$; 2) The representation should discard image details that are irrelevant for text prediction to obtain a correspondence of text and image components.

To address the former, we train a text decoder $p(\rw|\rz)$ on top of the image representation. This results in the full graphical model depicted in Figure~\ref{fig:graph}. According to the graphical model, we can now define the joint distribution of our data $p(\rx, \rw) = p(\rw|\rx) p(\rx)$, where we are particularly interested in the conditional
\begin{align}
    p(&\rw | \rx) = \prod_i p(\rw_i|\rx)\\
        &= \prod_i \iint p(\rw_i|\rz) p(\rz|\rx,\rc) p(\rc|\rx) d\rc d\rz\\
        &= \prod_i \iint p(\rw_i|\rz) \sum_k \pi^{(k)}(\rx) \mathcal{N}\Big(z|\mu(\rx^{c^{(k)}}), \sigma(\rx^{c^{(k)}})\Big)\\
        &\qquad\qquad\qquad\qquad\qquad p(\rc^{(k)}|\rx) d\rc^{(k)} d\rz\\
        &= \prod_i \sum_k \pi^{(k)}(\rx) \E_{\rz\sim \mathcal{N}(z|\mu(\rx^{c^{(k)}}), \sigma(\rx^{c^{(k)}}))}\Big[p(\rw_i|\rz)\Big]
\end{align}
where the crop parameters $\rc^{(k)}$ are deterministically determined as a parametric function of $\rx$, which is represented by the transformer architecture.

By parameterizing the conditional distribution with our model parameters, we can define our objective to minimize the negative log-likelihood of the joint occurrence of pictures $\rx$ and text $\rw$:

\begin{align}
    \min_{\theta}&\E_{\rx,\rw\sim \tilde{p}}\Big[-\log p_{\theta}(\rx, \rw)\Big] \\
        &= H(\rx) + \min_{\theta}\E_{\rx,\rw\sim \tilde{p}}\Big[-\log p_{\theta}(\rw | \rx)\Big]
\end{align}

where $H(\rx)$ is the entropy of the image data and as a constant is irrelevant for the optimization procedure. Specifically, the second term on the right-hand side is the negative log-likelihood of the text data given the image under our model and will be denoted as NLL. By minimizing NLL, we are effectively capturing the information required to predict $\rw$ since the predictive text distribution $p(\rw|\rz)$ depends only on the representation $\rz$.

Since we want the representation $\rz$ to focus only on the image details which have a textual counterpart, the second condition aims to discard any image-specific detail from $\rz$. This can be done by minimizing  the distance between the conditional distributions $p(\rz|\rx)$ and $p(\rz|\rw)$ to enforce consistency between encoded images and corresponding text.
 While we model $p(\rz|\rx)$ directly, $p(\rz|\rw)$ is intractable under our graphical model.
Hence, we introduce a variational distribution $q(\rz|\rw)$ to approximate the representation induced by the observation of the text data. We model $q(\rz|\rw)$ as a Gaussian mixture distribution in which each component corresponds to a single word to capture the compositional nature of the text. 
By minimizing the Kullback-Leibler (KL) divergence between $p(\rz|\rx)$ and $q(\rz|\rw)$ we are effectively minimizing the amount of image-specific information embedded into $\rz$, fulfilling our second requirement:
\begin{align}
    &\KL \Big( p_{\theta}(\rz|\rx) \big\Vert q_{\phi}(\rz|\rw) \Big) \ge \KL \Big( p(\rz|\rx) \big\Vert p(\rz|\rw)\Big)\\
    &= \KL \Big( p(\rz|\rx,\rw) \big\Vert p(\rz|\rw)\Big) \\
    &= I(\rz; \rx | \rw).
\end{align}
When $I(\rz; \rx | \rw)$ is minimal, the representation $\rvz$ must contain only the information that can be determined through the text (as $\rz$ and $\rx$ become conditionally independent), discarding picture specific nuisances not mentioned in the textual description $\rw$.

Combining the KL-divergence term with NLL, we arrive at the loss function
\begin{align}
    \min_{\theta,\phi} \mathcal{L} = \min_{\theta,\phi} E_{\rx,\rw\sim \tilde{p}}\Big[&-\log p_{\theta}(\rw | \rx)\\
    &+ \KL \Big( p_{\theta}(\rz|\rx) \big\Vert q_{\phi}(\rz|\rw) \Big)\Big].
\end{align}
The KL-divergence between the two mixture models $p(\rz|\rx)$ and $q(\rz|\rw)$ is approximated by sampling from each Gaussian component using the reparameterization trick~\cite{kingma2014} resulting in a stochastic estimate of the loss.

\subsection{Image Area Loss}
In the previous sections, we provided a detailed description of how a compositional Gaussian mixture representation for both text and image may be learned. Learning the representation is the key part that makes CoMix implicitly split the image into elements that match the textual components.
However, the crops that are learned by the spatial transformer are not guaranteed to be minimally enclosing the objects they represent. Since neural networks are able to learn arbitrary mappings from the image region to representation, the network can learn to attend to a larger image part than just the object, effectively including unnecessary information without harming the modeling performance.

We are specifically interested in the smallest bounding box per object to be encoded into a Gaussian component because a precise localization greatly helps model inspection and interpretability.
Hence, we introduce another tunable loss term that penalizes the size of the image crops learned by the spatial transformer.
Our final loss is given by
\begin{align}
    \min_{\theta,\phi} \mathcal{L} = \min_{\theta,\phi} E_{\rx,\rw\sim \tilde{p}}\Big[&-\log p_{\theta}(\rw | \rx)\\
    &+ \KL \Big( p_{\theta}(\rz|\rx) \big\Vert q_{\phi}(\rz|\rw) \Big)\\
    &+ \lambda \big|\prod_i A_{\theta,ii}\big|\Big].
\end{align}
where the product of the diagonal elements of the affine transformation matrix $A_{\theta}$ equal the area of the patches by the transformer architecture. The coefficient $\lambda$ regulates to which degree we would like to penalize the area of the image crops. In the experiments section below, we investigate the effect of this hyperparameter.

\section{Experiments}

In this section, we describe the datasets and experimental tasks and provide quantitative and qualitative results evaluating the representation learned by CoMix. Furthermore, we demonstrate the performance of our model on classification, image-retrieval, and weakly supervised object detection.

\subsection{Datasets and Setup}

\myparagraph{Datasets.} We conduct experiments on two datasets, MultiMNIST and MultiCIFAR10, which represent variations of MNIST~\cite{mnist98} and CIFAR-10~\cite{Krizhevsky09learningmultiple}, respectively. MNIST consists of 60K/10K training/test examples from 10 handwritten digits and CIFAR-10 contains 50K/10K training/test natural image examples from 10 classes.

In contrast, MultiMNIST and MultiCIFAR10 combine several original images randomly sampled onto the same canvas. For each example, we sample between one to four original images, scale them with a factor drawn from $\mathcal{N}(1.3, 0.1)$, and place them at random locations on the canvas. The canvas size is $56 \times 56$ for MultiMNIST and $80 \times 80$ for MultiCIFAR10. The presence of an object class in the image is indicated with the help of labels, regardless of how many times the same object appears. We sample training and test data once according to the original data sizes and then keep the datasets fixed across all the experiments.
See Figure~\ref{fig:qual} for the examples of images from these datasets.

\myparagraph{Experimental setting.} All the compared models consist of the same neural-network architectures with the similar or same number of parameters. For MultiMNIST, the image encoder is a 3-layer convolutional network with BatchNorm and ReLU after each layer. For MultiCIFAR10, we use ResNet18 as the image encoder. For both datasets, we use an embedding layer for the text encoder, a 2-layer MLP with ReLUs as the text decoder, and a spatial transformer network consisting of a 3-layer convolutional network with BatchNorm and ReLU.

The number of the mixture components $k$ learned by the spatial transformer network is set to 5 across all the datasets such that the model has to learn to actively choose to use or not to use components. The decision about using a particular component is regulated by the mixture weights $\pi^{(k)}$, which is a learned output of the same network. The area loss coefficient $\lambda$ is set to 4 for both datasets (see Section~\ref{sec:evalarealoss}).
We randomly split 10\% of the training data as a validation set to tune the remaining hyperparameters.
The model's code, experiments, and data generation will be made publicly available.

\subsection{Classification and Object Detection}

\begin{table}
    \centering
    \begin{tabular}{l c c c c}
         & \multicolumn{2}{c}{MultiMNIST} & \multicolumn{2}{c}{MultiCIFAR10} \\
         \cmidrule(lr){2-3}\cmidrule(lr){4-5}
         & Cls & Det & Cls & Det \\
        \midrule
        CNN  & 95.87 & n/a & 73.99 & n/a\\
        WSDDN  & \textbf{99.63} & \textbf{88.27} & 75.42 & 49.47\\
        CoMix & \textbf{99.39} & \textbf{87.04} & \textbf{90.74} & \textbf{75.83} \\
        \midrule
    \end{tabular}%
    \caption{Classification and object detection performance measured in mean average precision (mAP, in percent) on MultiMNIST and MultiCIFAR10 datasets.}
    \label{tab:wsod_results}
\end{table}

\myparagraph{Task.} We evaluate our model on its ability to detect and classify objects in an image. For the classification, since there are multiple target classes per image, we measure the mean average precision (mAP) of the predictions of our model's text decoder. 
While our model has a direct supervision on the classification task, there is no supervision signal or loss on object detection. Solely the learning of a compositional representation as well as the spatial-transformer architecture facilitates the fact that our model naturally learns to detect objects without the ground-truth bounding-box supervision. Thus, the task setup is identical to weakly supervised object detection (WSOD).

The predicted bounding boxes come from the transformer networks predictions that select a region of the image to be used as one component in our mixture model. We evaluate on the ground-truth bounding boxes that are known from the data generation process, i.e., location and boundary of each individual original image. Following \cite{pascal-voc-2012}, object detection is measured by mAP, where a detection is considered to be a true positive whenever the intersection over union (IOU) of the predicted and the ground-truth bounding box is greater than 0.5.

\myparagraph{Baselines.} Two baselines are introduced for the classification and object-detection tasks. Firstly, we isolate the image encoder of our model to do the classification directly from the input image to the label output without any representation learning denoted as CNN. This baseline is exclusively trained on a binary cross-entropy loss to predict the presence of the object classes in the image. Our model should match the CNNs classification performance to ensure that the representation learning does not hinder the prediction performance.

Secondly, we compare against Weakly Supervised Deep Detection Network (WSDDN)~\cite{BilenV16} serving as an object-detection baseline. WSDDN is an established backbone for the weakly supervised object-detection networks that relies on adaptive pooling over a convolutional feature map similar to modern supervised object detection networks such as Faster-RCNN~\cite{RenHGS15}. Contrary to our approach, bounding-box proposals are not learned, but generated algorithmically using the selective windows search strategy~\cite{SandeUGS11}.

\begin{figure}
  \centering
  \includegraphics[width=0.95\linewidth]{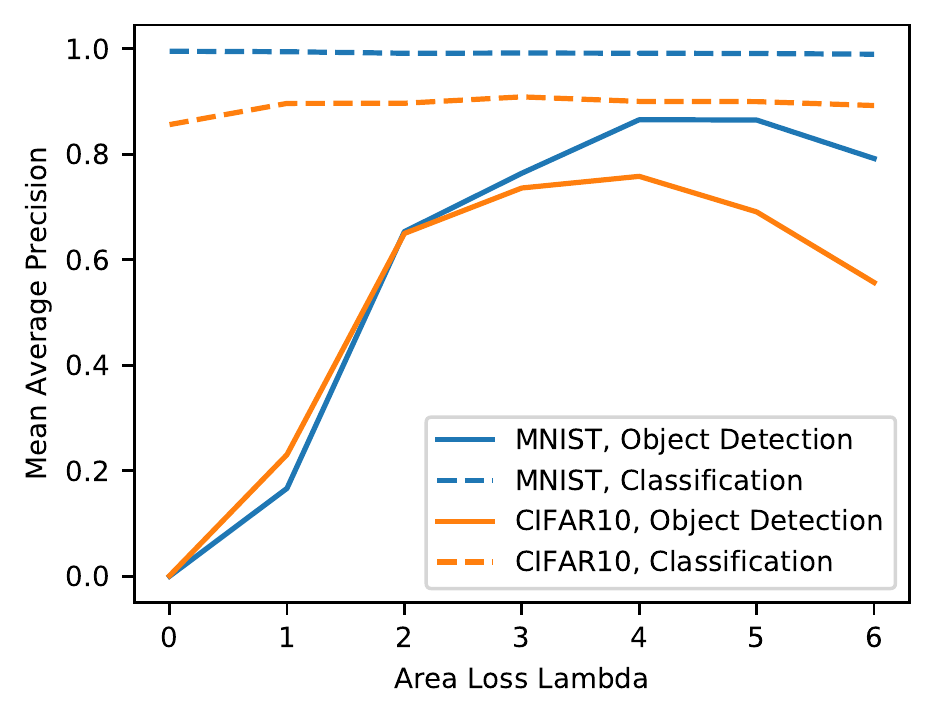}
\caption{Mean average precision of our CoMix model on both classification as well as object detection with varying values of $\lambda$. The hyperparameter $\lambda$ enables an area loss term that tightens bounding box predictions around objects.}
\label{fig:arealoss}
\end{figure}

\begin{table*}
    \centering
    \begin{tabular}{l c c c c c c c c}
         & \multicolumn{4}{c}{MultiMNIST} & \multicolumn{4}{c}{MultiCIFAR10} \\
        \cmidrule(lr){2-5}\cmidrule(lr){6-9}
         & avg. Rank & R@1 & R@5 & R@10 & avg. Rank & R@1 & R@5 & R@10\\
        \midrule
        \textit{Regular} &&&&&&&&\\
        Gauss & 6.85 & 21.09 & 64.84 & 86.72 & 16.47 & 14.06 & 32.03 & 55.47\\
        CoMix & \textbf{2.33} & \textbf{64.06} & \textbf{91.41} & \textbf{96.88} & \textbf{2.5} & \textbf{46.09} & \textbf{93.75} & \textbf{97.66} \vspace{2mm}\\

        \textit{Skewed} &&&&&&&&\\
        Gauss & 15.96 & 31.25 & 40.63 & 53.13 & 22.35 & 7.81 & 35.16 & 49.22\\
        CoMix & \textbf{2.45} & \textbf{57.03} & \textbf{89.84} & \textbf{98.44} & \textbf{5.45} & \textbf{47.66} & \textbf{77.34} & \textbf{85.94} \\

        \midrule

    \end{tabular}%
    \caption{Image retrieval results of our CoMix model compared to the Gauss baseline where we replace our composition Gaussian mixture latent representation with a single Gaussian distribution. \textit{Regular} refers to the normal version of our datasets MultiMNIST and MultiCIFAR10, whereas \textit{skewed} indicates a more difficult version where the combination of seen objects is highly correlated during training. Scores refer to the average retrieval rank and recall percentage at rank 1, 5 and 10.}
    \label{tab:retr_results}
\end{table*}

\myparagraph{Results.}
We report the classification and object detection results in Table~\ref{tab:wsod_results}. On classification, our CoMix model outperforms both baselines on MultiCIFAR10 by a large margin (90.7\% vs 75.4\%/74\%) and is on par with WSDDN on MNIST. The close results on MultiMNIST can most likely be explained by getting close to the perfect classification results rather than choices in the model. We believe, the considerably higher performance of our model on MultiCIFAR10 can be attributed to the compositional modelling of the individual objects in the image. For this reason, mAP approaches ResNet18's classification performance of around 93\% on the individual CIFAR10 images. Since our model processes image parts separately, we can generalize from the single image CIFAR10 classification performance to our more difficult MultiCIFAR10 dataset. Both baseline models back this property and, therefore, fall short in classification.

Similarly on the object detection, CoMix obtains a comparable performance as WSDDN on MultiMNIST, but outperforms it on MultiCIFAR10 (75.8\% vs. 49.5\%). Being able to flexibly learn the location of the objects instead of relying on the statically generated bounding boxes helps our model to more accurately pinpoint the image region responsible for a class label. Both results show the benefit of modelling image parts in isolation instead of the whole image as once (CNN) and of learning bounding boxes for our components with the spatial transformer module.

\subsection{Inspecting the Area Loss}
\label{sec:evalarealoss}

We argue that our area loss is essential for our model to learn bounding boxes that enclose the object tightly, which is important for both the detection performance and model inspection. In order to validate this claim, we train our model on both datasets with varying values of $\lambda$. A $\lambda$ of 0 indicates that no area-loss term was applied while the bigger $\lambda$ gets, the more bounding boxes will be penalized for their size. In Figure~\ref{fig:arealoss}, we report mAP for both the classification and object detection for different values of $\lambda$. Interestingly, the classification performance is largely unaffected with increasing $\lambda$ and it is even higher for MultiCIFAR10 for any $\lambda > 0$ compared to no area loss.

The mAP of the weakly supervised object-detection task steadily increases for both datasets until $\lambda$ reaches a value of 4. A value greater than 4 enforces bounding boxes that are too small and, thus, detection accuracy diminishes. Based on this analysis, we set $\lambda = 4$ for our experiments. Most importantly, introducing this loss term only benefits both the detection and classification in the range that we tested, so that its use can be easily justified.

\subsection{Image Retrieval}

\myparagraph{Task.} To evaluate our model's representation learning capabilities, we introduce an image-retrieval task. Given a text sample, the task is to find the best matching image from a set of images. If our model learns a good shared representation of image-text pairs, it should be possible to retrieve the image matching a query text by finding the image representation closest to the inferred text representation. With the use of our model, we first encode both the text and all the images into Gaussian mixtures independently. Then, we score the text representation to all the image representations. Since these representations are multi-modal distributions, we resort to calculating the piece-wise distance between Gaussian components of all the possible text-image pairs. The distance is defined by the average euclidean distance of each text component's mean to the closest image component's mean.

Based on the distance score, we can rank the images from the high to low similarity. To evaluate retrieval performance, we report the average rank of the ground-truth matching image as well as the recall at ranks 1, 5, and 10.

\myparagraph{Ablation Study.} A key advantage of our model is that its representations are compositional. To study the impact of having a compositional representation, we create an ablation model (Gauss), where the latent variable $\rz$ is a Gaussian instead of a mixture of Gaussians. In this ablation model, the image encoder directly infers the latent variable $\rz$ from the input image without the compositional spatial transformer. Moreover, the text encoder's embedding layer is replaced by a 2-layer MLP such that the representation of the full sentence can be learned as it can no longer be composed of simple embeddings. Since the ablation model only uses a single Gaussian, the retrieval distance is calculated by taking the euclidean distance between the means of the text and image latents.

\myparagraph{Skewed Datasets.} To better contrast two different representation-learning approaches, we conduct experiments on a skewed version of MultiMNIST and MultiCIFAR10. During the training, there are three groups of object classes, where each object only co-occurs with objects from the same group. For instance, for MultiMNIST, only the digits (0, 1, 2) would be seen together on the same canvas and similarly for digit groups (3, 4, 5) as well as (6, 7, 8, 9).

During the test phase, all the digits can co-occur with any other digits as in our initial MultiMNIST and MultiCIFAR10 definition. The purpose of this skewed dataset versions is to show how compositionality can help overcome dataset bias, where objects are highly correlated in the training data, but might occur in the novel combinations during the test time.

\begin{figure*}
\centering
\begin{subfigure}{.48\textwidth}
  \centering
  \includegraphics[width=0.95\linewidth]{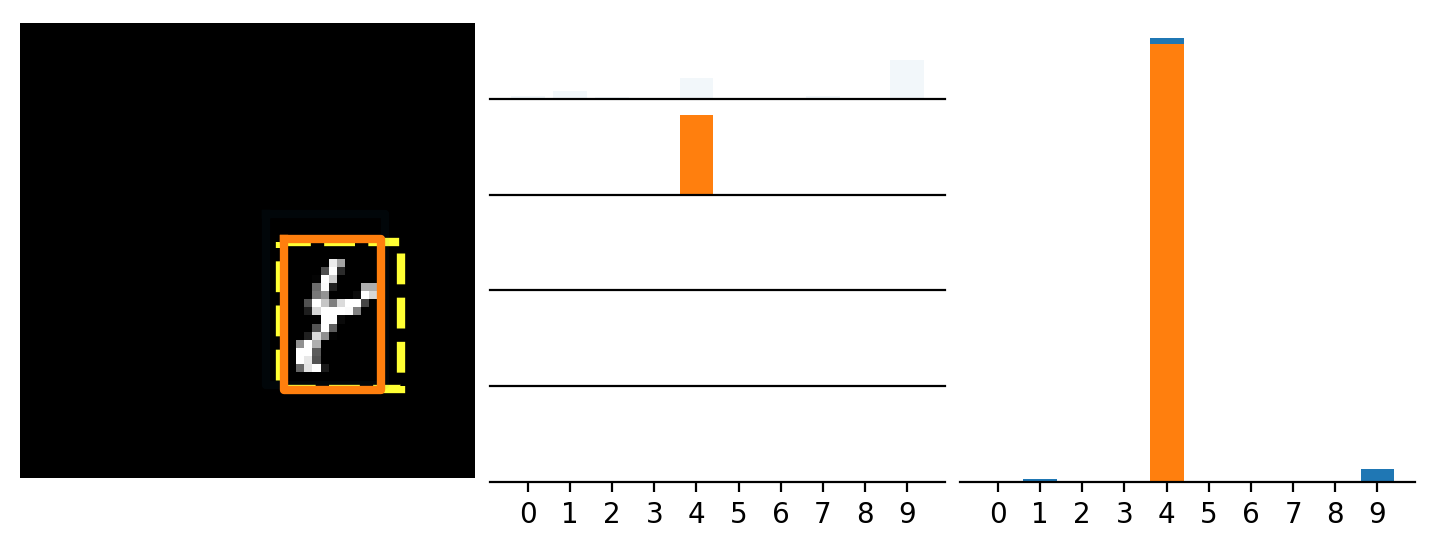}
\end{subfigure}%
\begin{subfigure}{.48\textwidth}
  \centering
  \includegraphics[width=0.95\linewidth]{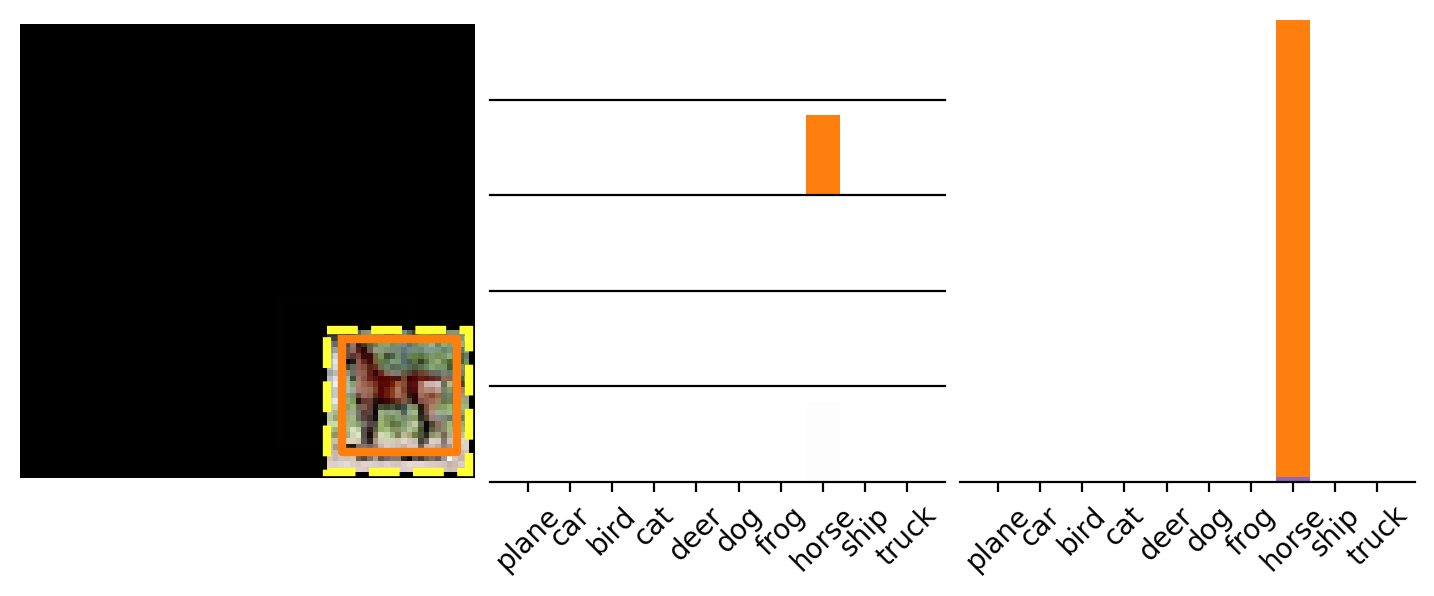}
\end{subfigure}
\begin{subfigure}{.48\textwidth}
  \centering
  \includegraphics[width=0.95\linewidth]{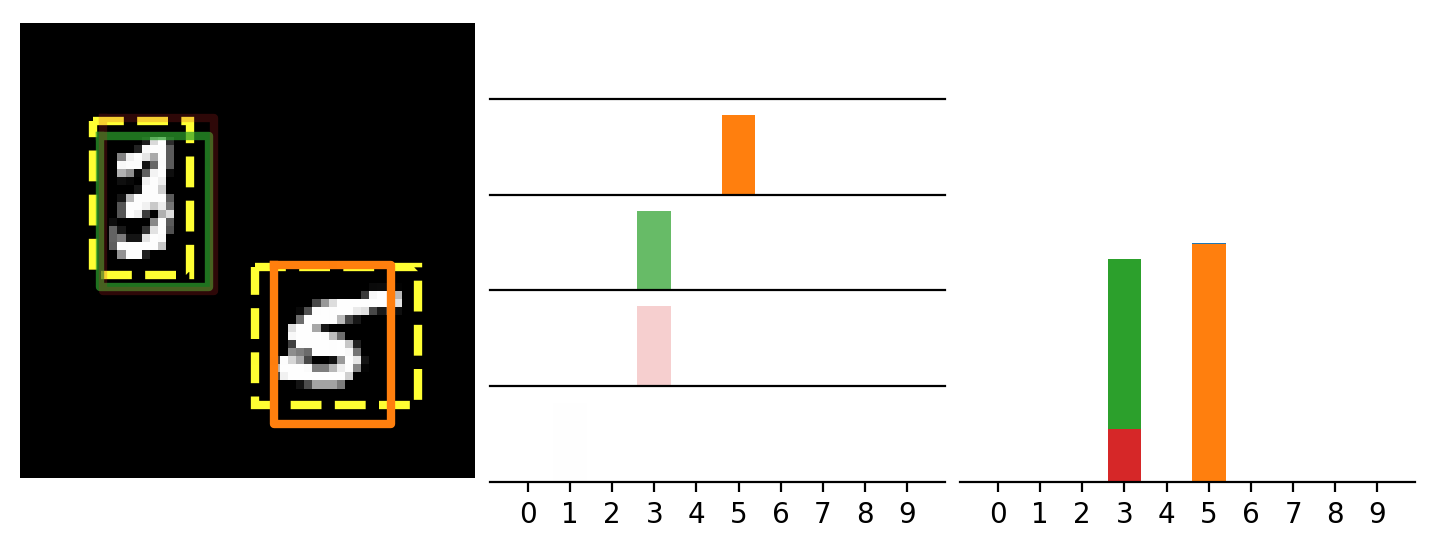}
\end{subfigure}%
\begin{subfigure}{.48\textwidth}
  \centering
  \includegraphics[width=0.95\linewidth]{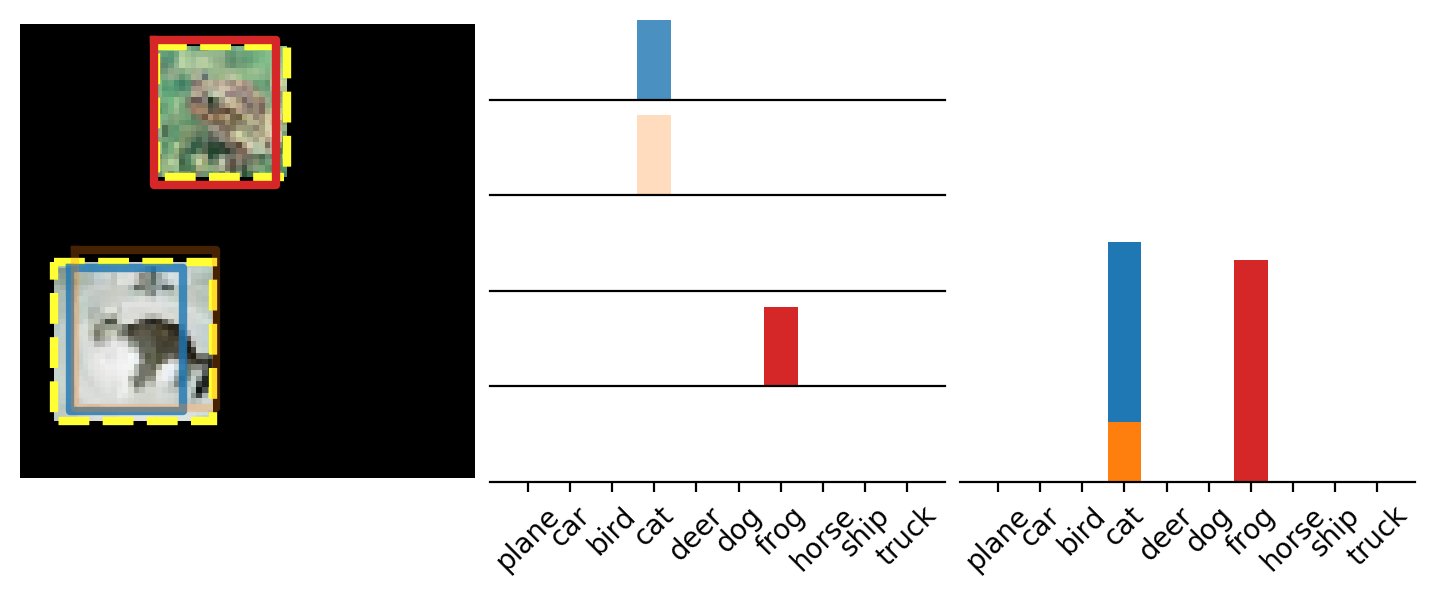}
\end{subfigure}
\begin{subfigure}{.48\textwidth}
  \centering
  \includegraphics[width=0.95\linewidth]{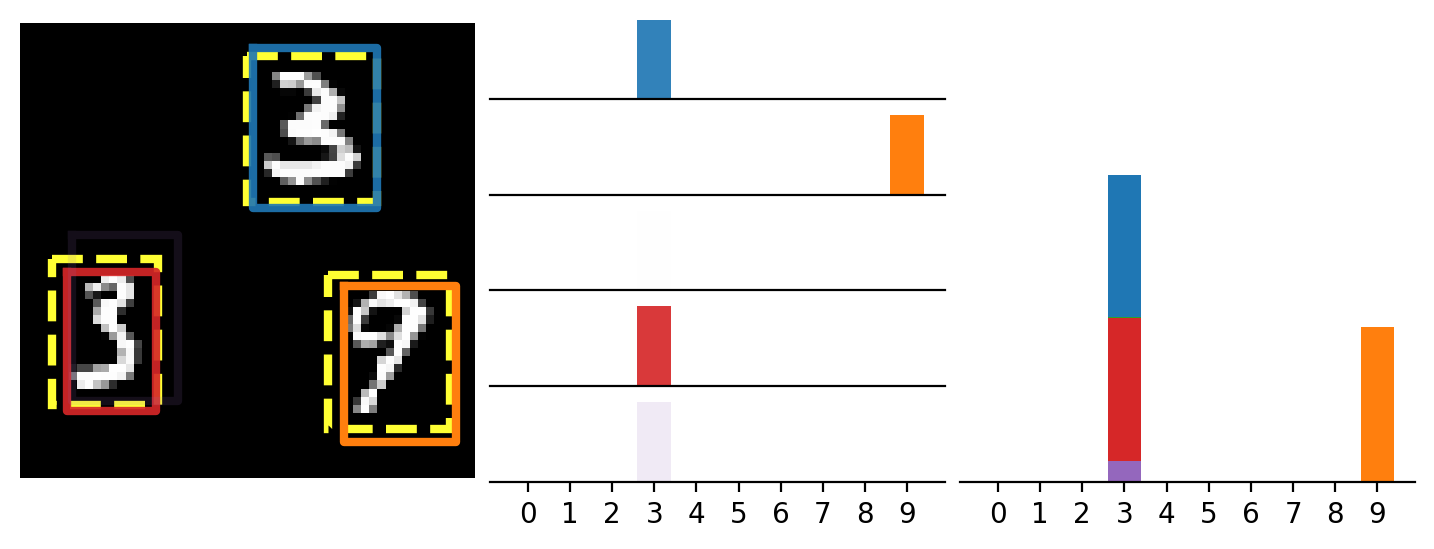}
\end{subfigure}%
\begin{subfigure}{.48\textwidth}
  \centering
  \includegraphics[width=0.95\linewidth]{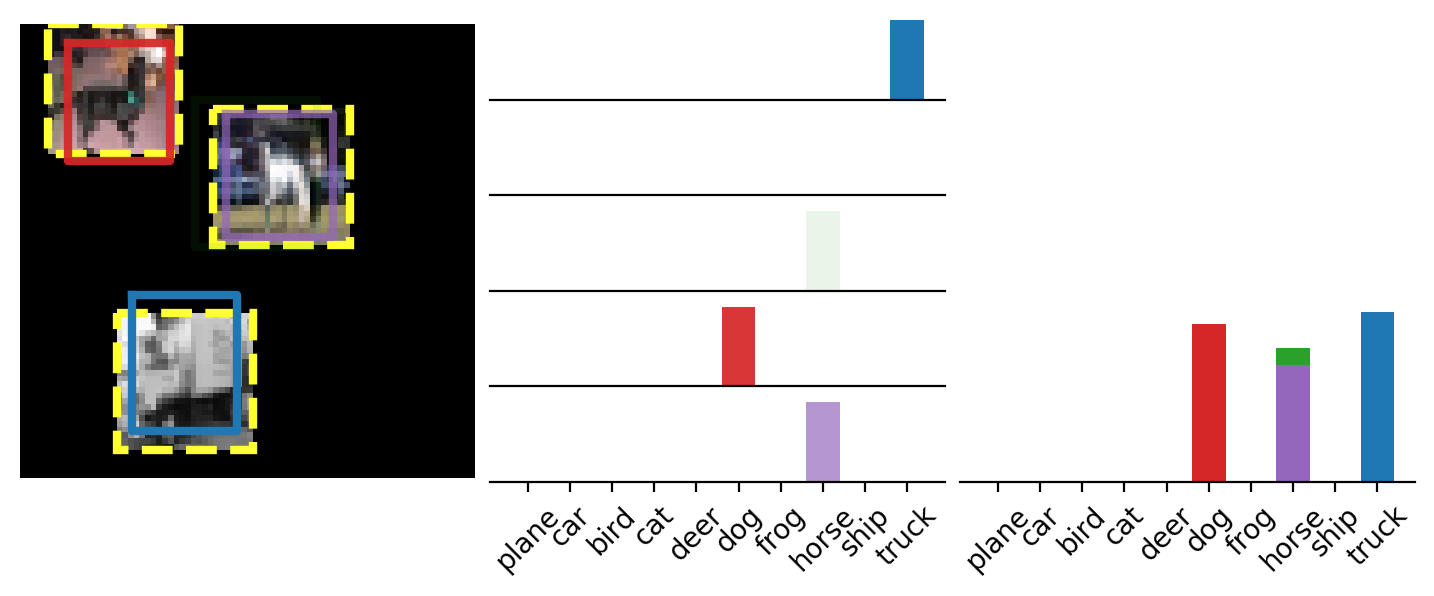}
\end{subfigure}
\begin{subfigure}{.48\textwidth}
  \centering
  \includegraphics[width=0.95\linewidth]{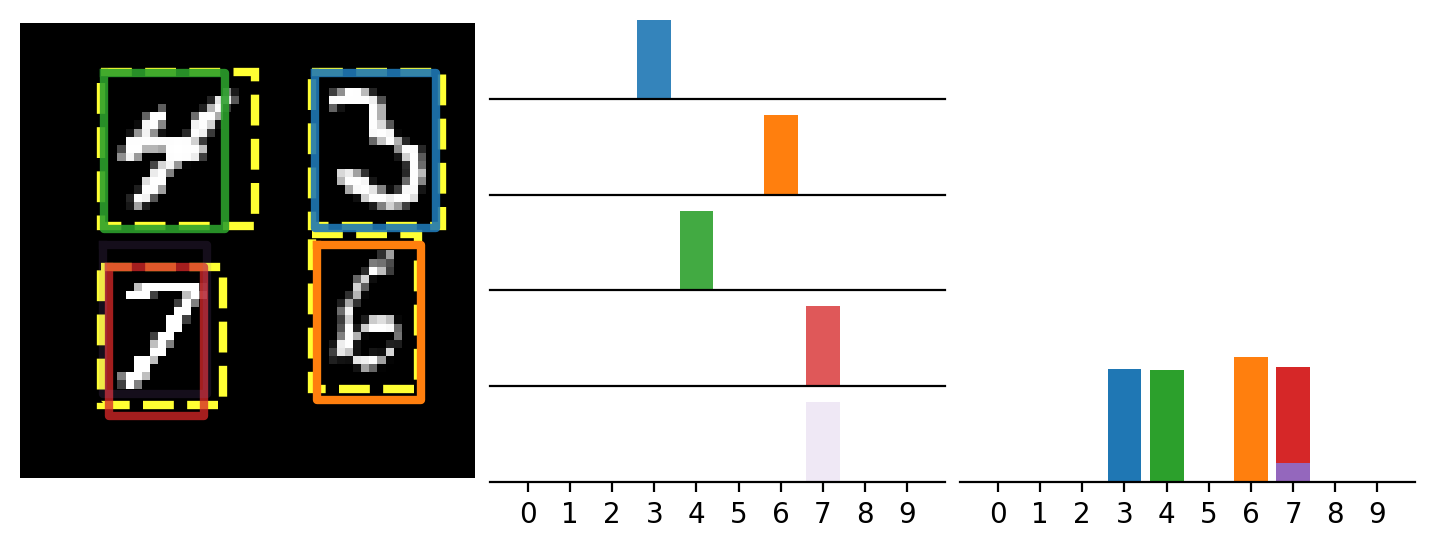}
\end{subfigure}%
\begin{subfigure}{.48\textwidth}
  \centering
  \includegraphics[width=0.95\linewidth]{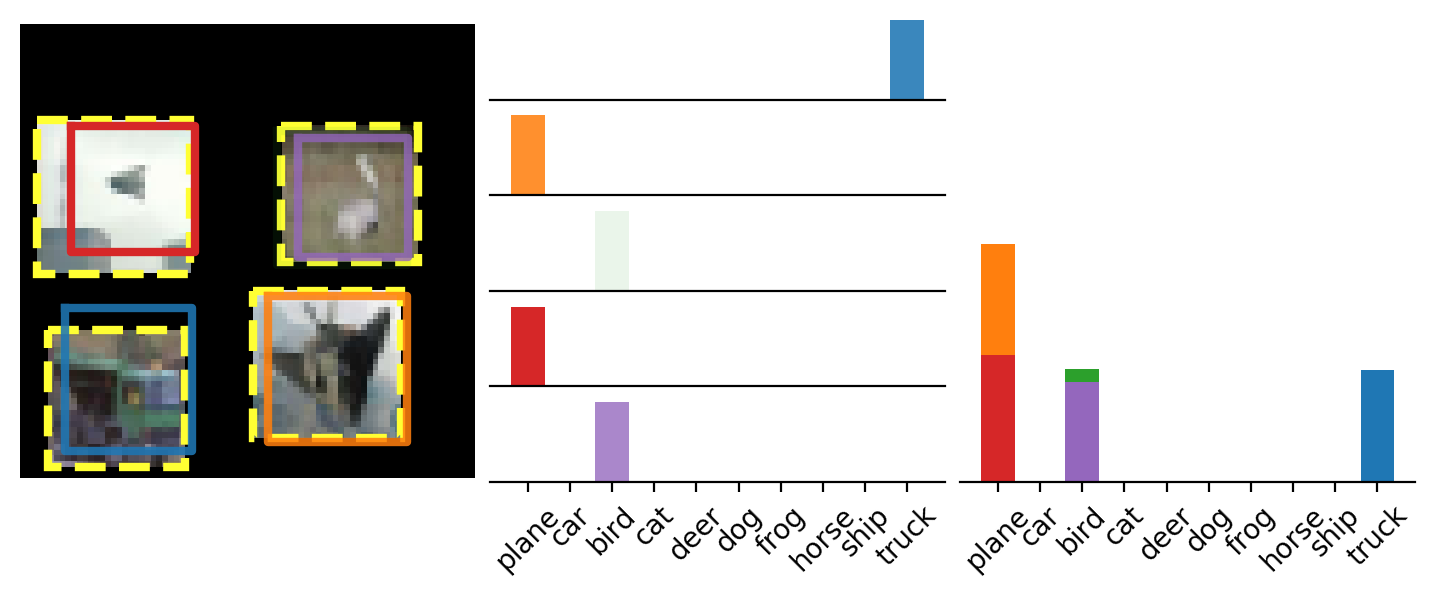}
\end{subfigure}
\caption{Qualitative results on MultiMNIST and MultiCIFAR10. For each dataset there are three columns: 1) data example visualizing the ground truth bounding boxes (yellow, dotted line) and the predicted bounding boxes, one for each component (various colors, solid line); 2) Categorical prediction of object for each individual component corresponding to bounding box with sample color; 3) Categorical mixture prediction of the whole image with colors indicating the contribution of each component.}
\label{fig:qual}
\vspace{-4mm}
\end{figure*}

\myparagraph{Results.}
Image-retrieval results are reported in Table~\ref{tab:retr_results}. We observe, learning a Gaussian mixture representation achieves a lower retrieval rank on average and a higher recall at all the measured levels as compared to a single Gaussian representation. There are two reasons for these results. Firstly, the Gaussian mixture of CoMix is multimodal and, thus, can model more complex latent distributions. Secondly, the compositionality of CoMix allows it to extrapolate to the unseen combination of objects during the test time. We observe this property when comparing the performance loss between the regular and skewed datasets. CoMix can maintain a low mean rank (2.3 $\rightarrow$2.5; 2.5 $\rightarrow$ 5.5), while average rank for the Gauss ablation gets considerably worse (6.9 $\rightarrow$ 16.0; 16.5 $\rightarrow$ 22.4) on both datasets.
In conclusion, the explicit modelling of a compositional representation greatly benefits CoMix's generalization and extrapolation performance of the downstream tasks.

\subsection{Qualitative Results}
To illustrate the datasets and inspect our model, we present qualitative example of CoMix applied to MultiMNIST and MultiCIFAR10 in Figure~\ref{fig:qual}. For each dataset, we show an example for 1 to 4 objects in the image. In each data image, we visualize the ground-truth bounding boxes of the objects (dashed yellow line) as well as the object detections by our CoMix model (blue, orange, green, red, purple boxes).

Each image is accompanied by two additional columns. The first column displays the textual prediction of the text decoder applied onto the representation of the patch with the same color. The opaqueness of the bounding boxes and the bar plots is proportional to the categorical mixture distribution $\pi$ that is predicted by the spatial transformer network. In other words, CoMix predicts the number of objects in the image through the $\pi$ values of each component and sets a components' $\pi^{(k)}$ (close) to zero when less than the maximum number of components are needed. The second additional column is the joint categorical class prediction of the whole image that is obtained by mixing the individual component prediction by their respective $\pi$ weights.

CoMix transparently shows which image patch causes which label prediction as visualized by the color in the last column. As an interesting side effect, the joint classification prediction reflects the count of objects in the image. For example, in the third row of the MultiMNIST data, the joint prediction probability of number 3 is twice the probability of number 9. Naturally, this is caused by the number 3 occurring twice in the image. Hence, our model learns to count without ever having received the supervision in this regard because the textual labels only indicate the presence of objects and not the number of them.

Being able to backtrack exactly how predictions are composed and caused by the concrete image regions is another strength of our model that makes it more interpretable. Due to the hard-attention mechanism of the spatial transformer module, CoMix only uses pixel information inside the bounding box to form its representations and the predictions. In contrast, while WSDDN also predicts bounding boxes, there is no exclusive relationship to the part of the picture the bounding box captures. The relationship is on a feature level that often has a receptive field covering the whole image.

\section{Conclusion}
We introduced our CoMix model that learns a compositional Gaussian mixture representation for both images and text. It utilizes the spatial-transformer architecture to decompose the raw image and expose transparently, which patches are responsible for the Gaussian components of the representation. The information content of the representation is guided by the textual data by employing a likelihood and KL-divergence loss. Without any additional supervision, CoMix learns to detect objects solely facilitated by an additional area loss.

We demonstrate the advantages of learning a compositional representation on the tasks of weakly supervised object detection and image retrieval, where we can validate that our model can generalize and extrapolate to an unseen combination of objects while, at the same time, being easier to inspect and interpret. Although our current results focus on synthetically generated datasets, a plausible next step would be to scale to the natural images with the natural-language captions, where the employment of recent language models such as BERT~\cite{DevlinCLT19}, GPT-3~\cite{BrownMRSKDNSSAA20} could enable richer textual representations. Taking further advantage from advances in contrastive learning such as in CLIP~\cite{RadfordKHRGASAM21}, our model could extend these approaches to be more interpretable and inherently exhibit compositionality in their representations.

\subsubsection*{Acknowledgments}
This work has been partially funded by the ERC (853489 - DEXIM) and by the DFG (2064/1 – Project number 390727645).

{\small
\bibliographystyle{ieee_fullname}
\bibliography{egbib}
}

\end{document}